\documentclass[a4paper,twoside]{article}

\usepackage{epsfig}
\usepackage{subcaption}
\usepackage{calc}
\usepackage{amssymb}
\usepackage{amstext}
\usepackage{amsmath}
\usepackage{amsthm}
\usepackage{multicol}
\usepackage{pslatex}
\usepackage{apalike}
\usepackage{algorithm2e}
\usepackage{booktabs} % Added to support your table styles (toprule/midrule)
\usepackage{graphicx} % Added to support \includegraphics
\usepackage{url}
\usepackage[bottom]{footmisc}
\usepackage{SCITEPRESS}

\begin{document}

\title{Adapting Multimodal Foundation Models for Few-Shot Learning: A Comprehensive Study on Contrastive Captioners}

\author{\authorname{Narasinghe N.K.B.M.P.K.B\sup{1} and Uthayasanker Thayasivam\sup{1}}
\affiliation{\sup{1}Department of Computer Science and Engineering, University of Moratuwa, Sri Lanka}
\email{patalee.21@cse.mrt.ac.lk, rtuthaya@cse.mrt.ac.lk}
}

\keywords{Few-Shot Learning, Multimodal Foundation Models, CoCa, Contrastive Learning, Parameter-Efficient Fine-Tuning, LoRA.}

\abstract{Large-scale multimodal foundation models, particularly Contrastive Captioners (CoCa), have achieved state-of-the-art results by unifying contrastive alignment with generative captioning. While zero-shot transfer capabilities are well-documented, the adaptation of these generative-contrastive hybrid models to downstream tasks with extreme data scarcity (few-shot learning) remains under-explored. Most of the current literature deals with dual encoders such as CLIP, leaving unanswered questions about how the distinct latent space of CoCa behaves under parameter-efficient fine-tuning (PEFT). In the paper, we set out a comprehensive empirical study on adapting the CoCa visual backbone for few-shot image classification. We systematically evaluate a hierarchy of strategies, ranging from training-free hybrid prototyping to deep parameter adaptation via Low-Rank Adaptation (LoRA). First, we identify an ``augmentation divergence'': even though strong data augmentation hurt a linear probe's performance in low-shot conditions, that is what is needed for stabilization of LoRA fine-tuning. We also demonstrate that hybrid objectives involving the use of Supervised Contrastive (SupCon) loss yield consistently better performance against normal Cross-Entropy across varying shot counts. Crucially, we characterize the sensitivity of training configurations are to data scarcity, which will serve as empirical reference settings for scaling regularization, rank, and sampling strategies to facilitate the efficient adaptation of generative-contrastive foundation models.}

\onecolumn \maketitle \normalsize \setcounter{footnote}{0} \vfill

\section{\uppercase{Introduction}}

The introduction of pre-trained foundation models on massive, diverse datasets has changed the face of deep learning. In vision-and-language, dual-encoder models like CLIP \cite{radford2021learningtransferablevisualmodels} and ALIGN \cite{jia2021scalingvisualvisionlanguagerepresentation} have established new benchmarks of zero-shot transfer learning by matching images and text in a common embedding space. Following this paradigm, the Contrastive Captioners (CoCa) model \cite{yu2022cocacontrastivecaptionersimagetext} combines contrastive and generative pre-training paradigms combined into one single and coherent design integrated with two objectives. This type of learning enables CoCa to acquire not only discriminative but also generative representations, demonstrating state-of-the-art performance across a wide variety of multimodal tasks.

Despite their powerful zero-shot capabilities, efficiently fine-tuning these large models on downstream tasks with limited labeled data,which is known as few-shot learning (FSL), remains extremely challenging. Full fine-tuning is often expensive in terms of computation resources, requires more data, and results in catastrophically forgetting the pre-training knowledge. Consequently, Parameter-Efficient Fine-Tuning (PEFT) methods, such as Low-Rank Adaptation (LoRA) \cite{hu2021loralowrankadaptationlarge}, have become a viable alternative by updating only a small amount of parameters.

Nonetheless, significant gaps exist in the current literature. Recent surveys confirmed that most of the few-shot adaptation works are focused to CLIP-based architectures, with emphasis on prompt learning methods such CoOp \cite{zhou2022learning} and cache-based adapters including Tip-Adapter \cite{zhang2022tip}.Whether such ``best practices'' that involve standard dual-encoders, could be applied into generative-contrastive hybrids such as CoCa remains an open question. More specifically, the interaction between deep parameter adaptation (LoRA), metric-based loss functions, and the intensity of data augmentation has yet to be evaluated systematically for that architecture.

This paper addresses these gaps by conducting a comprehensive empirical study on adapting the CoCa visual encoder. Moving beyond the prompt engineering focus of prior work, we explore complex adaptation techniques, forming a hierarchy of increasing complexity:
\begin{enumerate} 
    \item \textbf{Parameter-free hybrid prototyping} that fuses visual and textual embeddings without gradient updates. 
    \item \textbf{Linear probing} with a frozen visual encoder, analyzing the impact of augmentation intensity. 
    \item \textbf{LoRA fine-tuning} of the visual encoder utilizing adaptive configurations and hybrid objectives. 
\end{enumerate}

Our experiments on MiniImageNet yield three major contributions: first, we demonstrate that CoCa's pre-trained representations form a solid foundation for FSL, with simple prototype-based approaches achieving remarkable baselines. Second, we reveal an \textit{augmentation divergence}: strong augmentation harms linear probing in low data regimes, but is strictly necessary for LoRA convergence, avoiding overfitting. Finally, we show that mixed objectives with Supervised Contrastive (SupCon) loss consistently improve performance over vanilla Cross-Entropy and provide empirical reference setups to scale regularization and rank to help the efficient adaptation of generative-contrastive models.

\section{\uppercase{Related Work}}

\subsection{Multimodal Foundation Models}
The vision-and-language learning has gone from task-specific architectures into high scale foundation models. Single-encoder models mostly rule the roost on pure vision tasks, learning strong visual features but lacking inherent linguistic alignment. Dual-encoder models came onto the scene with CLIP \cite{radford2021learningtransferablevisualmodels} and ALIGN \cite{jia2021scalingvisualvisionlanguagerepresentation}, using contrastive learning to align visual and textual representations in a shared embedding space. This alignment enables powerful zero-shot capabilities for retrieval and classification.

Yet, dual-encoders often do not provide the granularity necessary to reason multimodally at a fine level. Generative models such as SimVLM \cite{wang2022simvlmsimplevisuallanguage} try to address this problem by training encoder-decoder architectures on prefix-language modeling tasks. Contrastive Captioners (CoCa) \cite{yu2022cocacontrastivecaptionersimagetext} merge these paradigms into a single framework. By combining an image encoder with a dual-purpose text decoder that includes a unimodal branch for contrastive alignment and a multimodal branch for generative captioning, CoCa achieves state-of-the-art performance across a broader range of tasks than dual-encoders alone. Unlike CLIP, which relies solely on global semantic alignment, CoCa's generative objective forces the visual encoder to capture fine-grained spatial details necessary for image captioning, making it a potentially superior backbone for few-shot adaptation.

\subsection{Few-Shot Adaptation Strategies}
Adapting foundation models to downstream tasks with limited data has emerged as an important research area. The first approaches focused primarily on \textbf{Prompt Learning}, inspired by NLP. Methods like CoOp \cite{zhou2022learning} and CoCoOp replace hand-crafted text prompts with learnable continuous vectors, optimizing the textual input while freezing the heavy image encoder. While effective, these methods are often limited by the expressivity of the text encoder and struggle to adapt the visual feature space itself.

\textbf{Adapter-based methods} constitute a different paradigm. Tip-Adapter \cite{zhang2022tip} proposes a training-free adaptation method by constructing a key-value cache of few-shot examples, blending pre-trained features with nearest-neighbor retrieval. While Tip-Adapter is highly efficient, it operates solely on the output features and cannot correct misalignment within the encoder's internal representations. Our work deviates from these ``frozen-backbone'' ones by employing deep parameter adaptation, which will enable the visual encoder itself to adjust to the target distribution.

\subsection{Parameter-Efficient Fine-Tuning (PEFT)}
Parameter-efficient fine-tuning has emerged as a standard through which deep adaptation might be achieved without the daunting expense of full fine-tuning. Full fine-tuning on smaller datasets is very resource intensive and, in fact, sometimes gives rise to catastrophic forgetting, where the model simply forgets the pre-trained general knowledge it possesses. PEFT mitigates this by updating only a small subset of parameters.

Classic techniques include \textbf{Adapter Modules} \cite{houlsby2019parameterefficienttransferlearningnlp}, which insert small bottleneck layers between transformer blocks, and \textbf{Prefix Tuning} \cite{li2021prefix}, which optimizes continuous prefixes for generation tasks. LoRA \textbf{Low-rank adaptation} \cite{hu2021loralowrankadaptationlarge} has been recently gaining a lot of traction because of its efficiency. LoRA hypothesizes that the weight updates during adaptation have a low ``intrinsic rank''. It injects trainable rank decomposition matrices into the transformer layers while freezing the pre-trained weights. Thus, it updates the model's behavior without adding parameters and adds no extra latency at inference time (if the weights can be merged). LoRA has been researched fully in the field of NLP (for instance, GPT-3); however, how it can be applied to visual backbones of generative-contrastive models like CoCa has not been studied sufficiently, particularly within the context of few-shot learning with a focus on metric-based regularizers.

\section{\uppercase{Methodology}}

\subsection{Data Preparation}
We utilize a balanced subset derived from the \textbf{Mini-ImageNet} benchmark \cite{vinyals2016matching}, which contains 100 classes. To construct a consistent low-data evaluation environment, we define fixed training and testing pools for all 100 classes. The training pool consists of the first 20 images from the original training split of each class, while the testing pool consists of the first 20 images from the original validation split. This results in a total of 4,000 images (2,000 for training, 2,000 for testing). The pipeline was implemented using the Hugging Face \texttt{datasets} library. For specific \(N\)-shot experiments (\(N \in \{1, 3, 5, 10, 20\}\)), we sample exactly \(N\) images per class from the training pool to create the adaptation set, while the entire testing pool is used for evaluation.

\subsection{Model and Adaptation Strategies}
We use a pre-trained CoCa model (ViT-L/14) with weights from \url{mscoco_finetuned_laion2B-s13B-b90k}. The model is kept frozen in evaluation mode for all experiments unless otherwise specified. Three adaptation strategies are explored.

\subsubsection{Strategy 1: Hybrid Prototype Classification}
This is a training-free method that leverages CoCa's multimodal nature.
\begin{itemize}
    \item \textbf{Visual Prototype:} For a class \(c\) with \(N\) support images \(\{I_1, ..., I_N\}\), the visual prototype is the normalized mean of their image embeddings.
    \[P_{image}(c) = \text{normalize}(\frac{1}{N} \sum_{i=1}^{N} f_{img}(I_i))\]
    \item \textbf{Textual Prototype:} We employ prompt ensembling over a fixed set of \(M\) templates (e.g., ``a photo of a \{class\}''). The textual prototype is computed as the normalized mean of the resulting embeddings:
    \[P_{text}(c) = \text{normalize}(\frac{1}{M} \sum_{j=1}^{M} f_{txt}(T_j))\]
    \item \textbf{Hybrid Fusion:} The final prototype is a weighted combination, where \(\alpha\) is a fusion hyperparameter.
    \[P_{hybrid}(c) = \text{normalize}((1-\alpha) P_{image}(c) + \alpha P_{text}(c))\]
\end{itemize}
A query image is classified by finding the class \(c\) with the highest cosine similarity between its embedding \(E_{query}\) and \(P_{hybrid}(c)\).

\subsubsection{Strategy 2: Linear Probing}
We attach a new linear classification head to the frozen CoCa image encoder. Only the parameters of this head are trained. Since the training set is small, it has been augmented with strong augmentation via the following strategies: Random Resized Crop, Horizontal Flip, Color Jitter, Random Grayscale, etc. The model is trained under AdamW \cite{loshchilov2017decoupled}, cross-entropy loss with label smoothing, and a cosine learning rate scheduler with warmup.

\subsubsection{Strategy 3: LoRA Fine-Tuning with Hybrid Objectives}
We employ Low-Rank Adaptation (LoRA) to adapt the internal weights of the frozen ViT image encoder. For a pre-trained weight matrix \(W_0 \in \mathbb{R}^{d \times k}\), LoRA constrains the update with a low-rank decomposition: \(h = W_0 x + \Delta W x = W_0 x + B A x\), where \(A \in \mathbb{R}^{r \times d}, B \in \mathbb{R}^{k \times r}\), and the rank \(r \ll \min(d, k)\). Only \(A\) and \(B\) are trained.

\begin{figure}[h]
    \centering
    \includegraphics[width=\columnwidth]{lora.png}
    \caption{\textbf{Overview of the Proposed LoRA Framework.} We inject low-rank adapters (A, B) into the frozen CoCa encoder. The model features a dual-head design: a classification head optimized via Cross-Entropy Loss and a projection head optimized via Supervised Contrastive (SupCon) Loss.}
    \label{fig:lora_arch}
\end{figure}

We utilize a \textbf{depth-specific LoRA configuration}, targeting specifically \texttt{mlp.c\_fc} (the first fully-connected layer in MLP blocks) and \texttt{mlp.c\_proj} (the projection layer in MLP blocks) within the later transformer blocks. The rank \(r\), scaling factor \(\alpha\), and the specific subset of target layers were tuned independently for each shot setting (3, 5, 10, 20) to maximize validation performance, allowing the model to balance between retaining pre-trained knowledge and adapting to sparse new data.

To enhance generalization, we implement a \textbf{dual-head architecture} (illustrated in Figure \ref{fig:lora_arch}) trained with a dynamic hybrid loss function:

\begin{equation}
\mathcal{L}_{total} = \mathcal{L}_{CE} + \lambda(t) \cdot \mathcal{L}_{SupCon}
\end{equation}

\noindent where \(\mathcal{L}_{CE}\) is the cross-entropy loss with label smoothing applied to a linear classification head. \(\lambda(t)\) is a time-dependent weighting factor that warms up over training epochs, allowing the classifier to stabilize before metric constraints are heavily enforced. \(\mathcal{L}_{SupCon}\) is Supervised Contrastive Loss \cite{khosla2020supervised} computed on a separate projection head (Linear-ReLU-Linear). 

\begin{equation}
\mathcal{L}_{SupCon} = \sum_{i \in I} \frac{-1}{|P(i)|} \sum_{p \in P(i)} \log \frac{\exp(z_i \cdot z_p / \tau)}{\sum_{a \in A(i)} \exp(z_i \cdot z_a / \tau)}
\label{eq:supcon}
\end{equation}

\noindent Here, \(z_i\) is the projected embedding, \(P(i)\) is the set of positive examples (same class) for anchor \(i\), and \(A(i)\) includes all samples in the batch. Crucially, we employ \textbf{temperature annealing} on the parameter \(\tau\). By decaying \(\tau\) from a higher initial value to a lower final value, we initially allow for looser alignment before forcing the model to learn tighter, more discriminative class boundaries in later epochs.

\section{\uppercase{Experiments and Results}}

\subsection{Experimental Settings}
% GLOBAL SETTINGS (Shared by everything)
We conduct all experiments on the \textbf{Mini-ImageNet} benchmark. We utilize the data splits and evaluation protocol detailed in Section III. All models were implemented in PyTorch and trained on a single NVIDIA P100 GPU. To ensure reproducibility, we fixed the random seed for all experiments.

\subsection{Experiment 1: Hybrid Prototype Classification}

\textit{Implementation Details:} We evaluated the text-visual fusion weight \(\alpha \in [0, 1]\). No gradient-based training was performed for this strategy; prototypes were computed directly from the pre-trained embeddings.

\begin{table}[h]
\caption{Accuracy (\%) of Hybrid Prototype Method across different text weights (\(\alpha\)).}
\label{tab:prototype}
\centering
\begin{tabular}{@{}c cccc@{}}
\toprule
\textbf{N-Shot} & \(\alpha = 0\) & \(\alpha = 0.2\) & \(\alpha = 0.5\) & \(\alpha = 0.7\) \\ 
\midrule
1  &  71.75   &  77.40  &  84.70   &  87.40   \\
3  &  86.10  &  88.45 &  91.05   &  90.50   \\
5  &  88.45   &  90.15   &  91.05   &  90.45   \\
10 &  90.30   &  91.60   &  91.95   & 90.90 \\
20 &  91.60   &  92.50   &  92.85   & 91.35 \\
\bottomrule
\end{tabular}
\end{table}

\textbf{Results:} Table \ref{tab:prototype} highlights three key findings. First, the hybrid fusion approach yields a substantial performance uplift over the visual-only baseline (\(\alpha=0\)) under all shot settings. This gain is most pronounced in the 1-shot regime, where the incorporation of textual priors (\(\alpha=0.7\)) increases accuracy by approximately 15.7\% (from 71.75\% to 87.40\%). Confirming that the rich text embeddings from CoCa effectively compensate for the high variance present in single-example visual prototypes.

Second, we observe a data-dependent shift in the optimal fusion weight. At extreme sparsity (\(N=1\)), a higher reliance on text (\(\alpha=0.7\)) is preferred; however, as the support set grows (\(N \ge 3\)), the optimal weight stabilizes at a balanced fusion (\(\alpha=0.5\)). Third, we observe diminishing marginal gains at higher shots: while the hybrid method maintains superiority at 20 shots (92.85\% vs. 91.60\%), the gap is not as large as that of 1-shot setting, suggesting that textual priors act primarily as a stabilizer for data scarcity. We further verified that these trends remain consistent across multiple random seeds, confirming that the hybrid robustness is not an artifact of specific image selection.

\subsection{Experiment 2: Linear Probing}

\textit{Implementation Details:} We trained a linear classification head on top of the frozen CoCa visual encoder using the AdamW optimizer. Preliminary experiments demonstrated that a cosine annealing scheduler (decaying from \(1 \times 10^{-3}\) to \(1 \times 10^{-6}\)) was critical for refining decision boundaries and extracting the final margin of fine-grained accuracy. Based on saturation trends, we fixed the training duration to 15 epochs. Given the susceptibility to overfitting in low-shot regimes, we performed a manual hyperparameter search for each shot setting, tuning the batch size, label smoothing, weight decay, and dropout rates to maximize validation accuracy. We compare three augmentation levels: \textit{None} (standard center crop), \textit{Low} (random resized crop and horizontal flip), and \textit{High} (adding color jitter and random grayscale).

\begin{table}[h]
\caption{Linear Probing Accuracy under Different Augmentation Levels.}
\label{tab:linear}
\centering
\begin{tabular}{@{}c c c c@{}}
\toprule
\textbf{N-Shot} & \textbf{No Aug} & \textbf{Low Aug} & \textbf{High Aug} \\ \midrule
1  & 67.35 & 72.10 & 67.65 \\
3  & 87.45 & 87.40 & 86.75 \\ 
5  & 91.40 & 90.50 & 88.65 \\
10  & 92.00 & 92.45 & 92.10 \\ 
20  & 93.25 & 93.10 & 92.70 \\
\bottomrule
\end{tabular}
\end{table}

\textbf{Results:} 
The results reveal distinct patterns in how augmentation affects performance across different data regimes. In the case of the extremely low data regime (1-shot), low augmentation provides a substantial benefit (+4.75 percentage points over no image augmentation), signifying that low data augmentation will help avoid overfitting when training data is severely limited. In high amounts of augmentation, on the other hand, shows minimal improvement, indicating that excessive augmentation may introduce too much variance for the model from which it cannot learn anything meaningful with a very few examples.

As training examples increase to 3-shot and then 5-shot, this advantage of augmentation is subsequently decreased, and then eventually reverses. At 3-shot, all augmentation levels perform similarly,  while with 5-shot, the highest accuracy (91.40\%) is achieved without augmentation, going ahead of low augmentation into the matter of 0.90 points and high augmentation by 2.75 points. This trend suggests that with sufficient examples, the learned representations benefit more from looking at the original data distribution than the augmented one.

Interestingly, at higher data regimes (10-shot and 20-shot), all augmentation methods performed similarly, with accuracy values converging around 92--93\%.  The narrow gap suggests that the classification head succeeds in adapting to the pre-trained encoder's representations when adequate training data is available, regardless of augmentation strategy.

\textbf{Hyperparameter Trends.} Table~\ref{tab:hyperparams} details the systematic variation in optimal training configurations. Here, we notice an interesting inverse relationship between the availability of data and the amount of regularization enforced. In the low-shot regime (1--3 shots), aggressive regularization was critical to prevent memorization of the sparse support set; optimal performance required small batch sizes (4--8), high dropout (0.3--0.5), and strong weight decay (0.1). On the contrary, when the support set grew larger (10--20 shots), these constraints could be relaxed significantly, allowing for larger batches (16--32) and reduced weight decay as the increased data diversity naturally stabilized the optimization of the linear head.

\begin{table}[h]
\caption{Optimal Hyperparameter Settings observed by Data Regime.}
\label{tab:hyperparams}
\centering
\resizebox{\columnwidth}{!}{%
\begin{tabular}{@{}l c c c c@{}}
\toprule
\textbf{N-Shot} & \textbf{Batch Size} & \textbf{Dropout} & \textbf{Weight Decay} & \textbf{Label Smooth.} \\ 
\midrule
1 & 4--8 & 0.5 & 0.1 & 0.2 \\
3 & 8 & 0.3--0.5 & 0.1 & 0.2 \\
5 & 8 & 0.3 & 0.1 & 0.2 \\
10 & 8--16 & 0.2--0.3 & 0.05--0.1 & 0.1--0.2 \\
20 & 16--32 & 0.2--0.3 & 0.05 & 0.1 \\
\bottomrule
\end{tabular}%
}
\end{table}

\subsection{Experiment 3: LoRA Fine-Tuning}

\textit{Implementation Details:} We fine-tuned the LoRA adapters using the AdamW optimizer. To ensure rigorous comparison, we performed an independent hyperparameter search for both the Cross-Entropy (CE) and Hybrid (CE + SupCon) objectives, wherein the epoch counts, weight decay, and dropout rates were tuned.

A major implementation difference arises with the sampling and scheduling of the data. For CE experiments, we used standard random sampling. However, for the Hybrid objective, standard sampling often yields batches lacking positive pairs (instances of the same class), thus rendering the contrastive loss ineffective. Therefore, we implemented a \textbf{stratified batch sampler} that constructs batches with a fixed number of instances per class to ensure proper contrastive computation.  Moreover, to ensure maximum stability, we used
\textbf{synchronized training schedules}:
\begin{itemize}
    \item \textbf{Learning Rate:} Cosine annealing learning rate scheduler for fine grained accuracy.
    \item \textbf{Temperature (\(\tau\)):} Annealed from high to low to progressively sharpen cluster boundaries.
    \item \textbf{Loss Weight (\(\lambda\)):} Dynamically scheduled (low \(\to\) high) to prioritize feature alignment in the mid-phase while allowing the classifier to refine decision boundaries at the start and end of training.
\end{itemize}

\begin{figure}[h]
    \centering
    \includegraphics[width=\columnwidth]{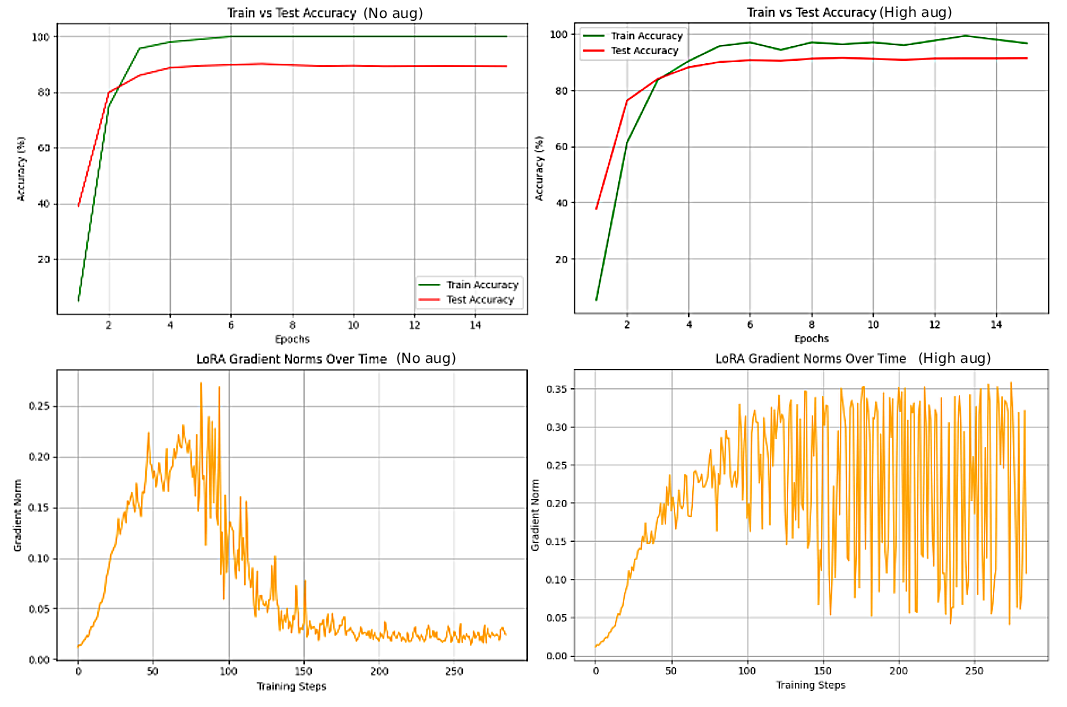}
    \caption{Training dynamics for 3-shot LoRA fine-tuning with and without data augmentation. Left: No augmentation. Right: Strong augmentation (TrivialAugment).}
    \label{fig:aug_comp}
\end{figure}

\textbf{Results:} \textit{Impact of Data Augmentation:} Unlike linear probing, where strong augmentation degraded performance, LoRA fine-tuning benefits substantially from aggressive augmented input in all shot settings. Figure \ref{fig:aug_comp} illustrates this effect: without augmentation (left), training accuracy rapidly reaches $100\%$ while test accuracy plateaus, with collapsing gradient norms signaling overfitting. Strong augmentation (right) maintained healthy gradient variance during training and yielded  $1$--$3$ percentage point improvements in test accuracy.

The reasons behind this divergence are rooted in the adaptation mechanisms of the methods. Linear probing trains only a shallow head atop frozen features; augmentation-induced variance cannot be accommodated by the fixed encoder and thus hinders the classifier. In contrast, LoRA adapts the encoder itself, enabling learning of representations robust to augmented inputs. The fact that gradient activity remains sustained under augmentation proves that the model is learning generalizable patterns rather than memorizing the support set.

\textit{Sampling Strategy Analysis:}Here, we saw that the choice of batch sampler strongly induces optimization stability according to the loss objective. For pure Cross-Entropy training, employing a stratified batch sampler (grouping class instances) has shown to harm generalization performance when compared to the plain random sampling. We attribute this to the high variance in gradients between batches, as each stratified batch represents a narrow slice of the class distribution. Consequently, we utilized \textbf{random sampling for the CE baseline} to ensure unbiased gradient estimation, while reserving \textbf{stratified sampling solely for the Hybrid objective}, where it is mathematically necessary to construct valid positive pairs for the contrastive loss.

\begin{table}[h]
\caption{LoRA Fine-Tuning Accuracy (N \(\ge\) 3).}
\label{tab:lora}
\centering
\begin{tabular}{@{}c c c@{}}
\toprule
\textbf{N-Shot} & \textbf{CE Loss} & \textbf{CE + SupCon } \\ \midrule
3  & 91.90 & 92.15 \\ 
5  & 93.15 & 93.10 \\
10 & 94.15 & 94.35 \\ 
20 & 95.10 & 95.25 \\
\bottomrule
\end{tabular}
\end{table}

\textit{Performance Analysis.} Table~\ref{tab:lora} demonstrates that LoRA fine-tuning consistently gives the highest classification accuracies in all data conditions, outperforming both the linear probing (Exp. 2) and the parameter-free hybrid prototype method (Exp. 1). As an example, in the extreme 3-shot setting, LoRA achieves $91.90\%$ accuracy with pure CE loss, surpassing linear probing ($87.45\%$) and hybrid prototypes ($91.05\%$). Accordingly, this performance edge scales robustly as data increases, with a peak accuracy of $95.25\%$ at 20-shots. The hybrid objective (CE + SupCon) provides modest but consistent gains over pure cross-entropy, particularly in higher shot settings, indicating that metric-based regularization offers complementary benefits to discriminative learning as the support set expands.

\textit{Configuration Trends.} The optimal LoRA configurations exhibit clear scaling patterns correlated with data availability. For low-data regimes (3-shot and 5-shot), a rank of $r=4$ ended up being sufficient in order to cover overfitting in the adaptation from the last 4 transformer blocks. However, convergence dynamics differed by objective: pure CE was almost always stabilized at within 15 epochs, whereas the hybrid loss required extended training (20 epochs) to balance the competing classification and clustering gradients.

As data availability increased (10-shot and 20-shot), the optimal capacity scaled to $r=8$ targeting 6--8 layers. Hybrid training within these bounds contained significantly longer training schedules (30 epochs) compared to CE (20 epochs), reflecting the increased optimization complexity. A critical distinction also emerged in batch sampling requirements: because the contrastive objective needed stratified sampling, larger logical batches (e.g., 24 for 3-shot: 8 classes \(\times\) 3 images)  were required to provide sufficiently large negative pairs, whilst CE optimization benefited from smaller, random batches (size 8--16) which provide regularizing gradient noise. The regularization parameters did not vary significantly across settings (dropout: $0.1$--$0.2$, weight decay: $0.05$), while the projection head dimension was scaled from 128 (low-shot) to 256 (high-shot) to accommodate the increased representational capacity required for larger support sets. Our synchronized scheduling strategy; progressively increasing the contrastive weight $\lambda$ ($0.05 \to 0.3$) while annealing the temperature $\tau$ ($0.2 \to 0.07$) was necessary to stabilize the optimization of multiple tasks and achieve peak performance.

\subsection{Analysis of Feature Space Adaptation}
\label{subsec:analysis}

\begin{figure}[h]
    \centering
    \includegraphics[width=\columnwidth]{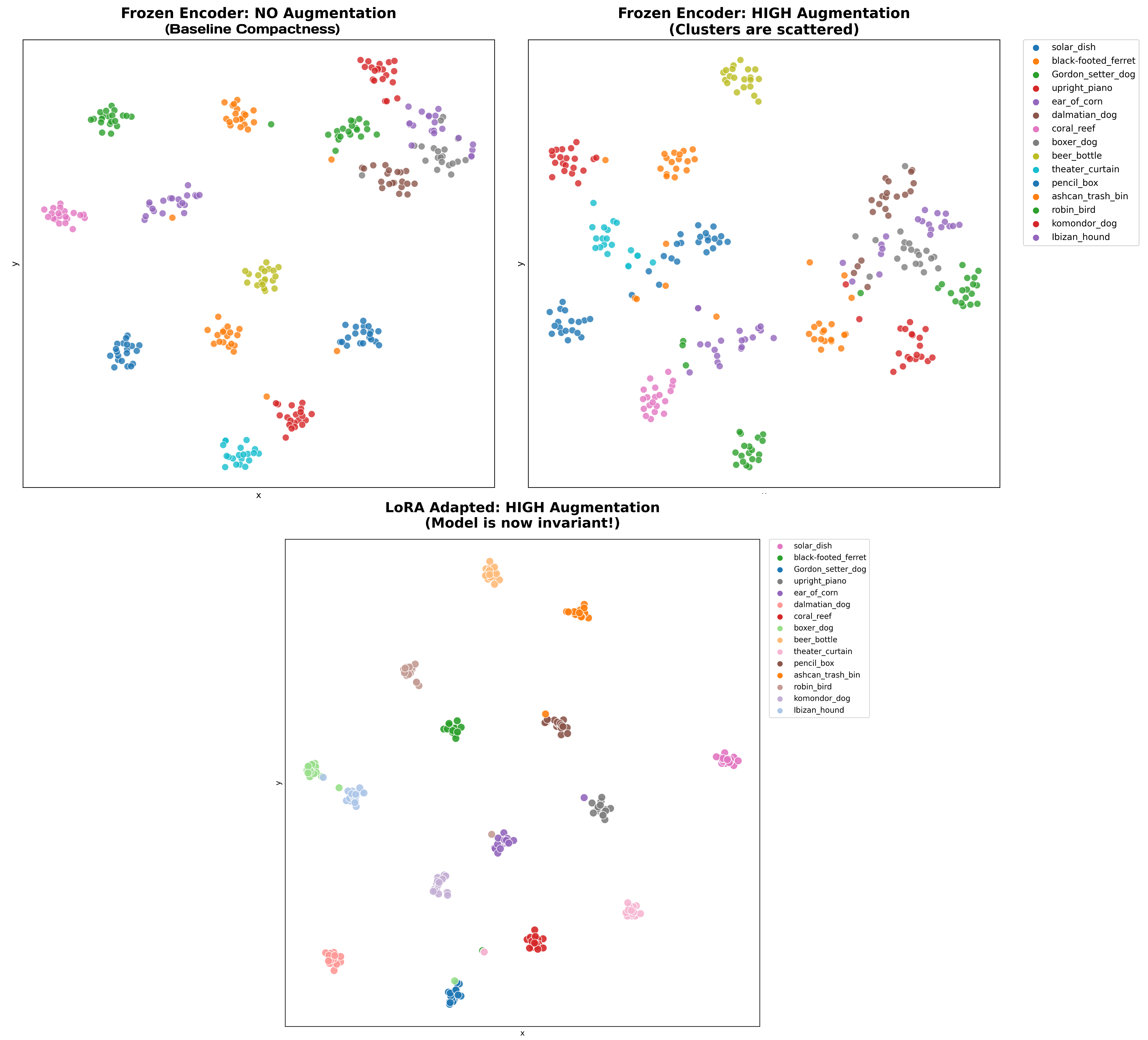}
    \caption{t-SNE visualization of feature embeddings for 15 randomly selected classes. \textbf{(Top-Left)} Frozen encoder with standard (clean) inputs shows compact clustering. \textbf{(Top-Right)} Frozen encoder with strong augmentation shows scattered clusters with high intra-class variance, explaining linear probe degradation. \textbf{(Bottom)} LoRA-adapted encoder with strong augmentation recovers cluster compactness, demonstrating learned invariance to distortions.}
    \label{fig:tsne_analysis}
\end{figure}

To validate the ``augmentation divergence'' hypothesis discussed in Section 4.3 and explain the robustness of the LoRA-adapted model, we visualize the feature manifolds using t-SNE \cite{vandermaaten2008tsne} projections in Figure \ref{fig:tsne_analysis}.

\textbf{Visualizing the Divergence.} 
The top-right plot in Figure \ref{fig:tsne_analysis} provides direct visual evidence of the fundamental challenge facing linear probing under strong augmentation. The resulting frozen feature clusters become very dispersed with points of the same class scattered over the embedding space. This spatial fragmentation directly reflects the high intra-class variance that makes the classification boundary intractable for a linear classifier. Visual inspection confirms that the performance drop is not because of the model's incapacity but because of the encoder's inability to maintain semantic cohesion when processing aggressively transformed inputs that lie outside its training distribution.

\textbf{Visualizing LoRA Invariance.} 
The bottom plot reveals how LoRA adaptation fundamentally restructures the feature manifold. Processing identical augmented inputs, the LoRA-tuned encoder produces markedly tighter, more cohesive clusters than even the clean baseline (Top-Left). This proves that LoRA is not only about restoring the quality of pre-trained representations. It instead learns augmentation-invariant transformations which compress intra-class variance while keeping inter-class separability intact. The enhanced cluster density shows that the low-rank adaptation successfully captures transformations introduced by augmentation creating a more robust feature space optimized for the downstream task.

\section{\uppercase{Conclusions and Future Work}}
\label{sec:conclusion}

This study presents a comprehensive empirical analysis of adapting the CoCa visual backbone for few-shot image classification. We create a clear hierarchy of adaptation methods indicating that while parameter-free \textbf{Hybrid Prototyping} is a strong baseline leveraging semantic text priors, \textbf{LoRA fine-tuning} remains requisite in achieving state-of-the-art results in low-data conditions.

A critical finding of this work is the \textbf{augmentation divergence} phenomenon: it was shown that while strong data augmentation degrades the performance of linear probing by introducing variance the frozen encoder cannot accommodate, it is necessary for stabilizing LoRA fine-tuning. Furthermore, our results confirm that integrating Supervised Contrastive (SupCon) loss with stratified sampling yields consistent gains over standard Cross-Entropy.

Looking forward, there are several promising paths that remain to be uncovered from the perspective of this research:

\begin{enumerate}
    \item \textbf{Attention-Aware Adaptation:} In this work, our LoRA configuration was restricted to the MLP layers (\texttt{c\_fc}, \texttt{c\_proj}) due to architectural constraints within the current OpenCLIP implementation. Future work should explore injecting adapters into the attention mechanism (Query, Key, and Value projections). Given the importance of attention dynamics in Vision Transformers, we hypothesize that adapting those layers may provide better parameter efficiency and fine control over the feature re-alignment.
    
    \item \textbf{Generative Few-Shot Learning:} Our study focused explicitly on the discriminative capabilities of the CoCa image encoder. However, CoCa's distinct advantage over dual-encoders (like CLIP) is its multimodal text decoder. Future research should investigate adapting the decoder for generative few-shot tasks, such as low-resource image captioning or visual question answering, to fully exploit the model's generative pre-training.
    
    \item \textbf{Holistic Multimodal Adaptation:}  While we applied adaptation to the visual backbone, the text encoder is frozen. Achieving co-adaptation of both modalities through coupled LoRA layers might provide even further alignment of latent spaces towards downstream tasks, potentially reducing the need for hand-engineered prompts in the hybrid prototyping stage.
\end{enumerate}

In summary, CoCa proves to be a highly adaptable foundation model. By carefully selecting between prototyping and fine-tuning based on the data regime, and by respecting the distinct augmentation needs of each strategy, practitioners can can intelligently harness these large-scale models within extremely constrained environments.

\bibliographystyle{apalike}
{\small
\bibliography{IEEEexample}}

\end{document}